\documentclass[11pt]{article}

\usepackage[preprint]{acl}

\usepackage{times}
\usepackage{latexsym}

\usepackage[T1]{fontenc}
\usepackage[utf8]{inputenc}

\usepackage{microtype}
\usepackage{amsmath}
\usepackage{amssymb}
\usepackage{inconsolata}

\usepackage{graphicx}
\usepackage[table,xcdraw]{xcolor}
\usepackage{booktabs}
\usepackage{multirow}
\usepackage[ruled]{algorithm2e}
\usepackage{enumitem}
\usepackage[most]{tcolorbox}
\DontPrintSemicolon
\SetKwInOut{KwIn}{Input}
\SetKwInOut{KwOut}{Output}

\newcommand{\ours}{Imagine-OPD}

\title{Thinking Without Images: Internalizing Visual Manipulation with On-Policy Self-Distillation}

\author{
  \textbf{Yishuo Cai\textsuperscript{1$*$}},
  \textbf{Jiahui Liu\textsuperscript{2$*$}},
  \textbf{Yuanxin Liu\textsuperscript{1}},
  \textbf{Haobo Deng\textsuperscript{2}},
\\
  \textbf{Linli Yao\textsuperscript{1}},
  \textbf{Yuhao Zheng\textsuperscript{3}},
  \textbf{Kun Ouyang\textsuperscript{1}},
  \textbf{Zhimo Li\textsuperscript{4}},
\\
  \textbf{Ziyue Wang\textsuperscript{1}},
  \textbf{Xu Sun\textsuperscript{1$\dagger$}},
  \textbf{Haoli Bai\textsuperscript{5}},
  \textbf{Xiaohui Li\textsuperscript{5}}
\\
\\
  \textsuperscript{1}State Key Laboratory of Multimedia Information Processing,\\
School of Computer Science, Peking University\\
  \textsuperscript{2}Central South University \quad
  \textsuperscript{3}University of Science and Technology of China\\
  \textsuperscript{4}Peking University \quad
  \textsuperscript{5}Huawei Technologies
\\
  \texttt{Emails: caiyishuo123@gmail.com} \quad
    \texttt{xusun@pku.edu.cn}
}

\begin{document}
\maketitle

\def\thefootnote{*}\footnotetext{Equal contribution. $^\dag$Corresponding authors.}\def\thefootnote{\arabic{footnote}}

\begin{abstract}
``Thinking with Images'' has emerged as an effective paradigm for fine-grained visual reasoning: by explicitly zooming into relevant regions and reasoning over crops, models can access local evidence that is difficult to recover from a single global image. However, this benefit comes with redundant tool invocations and longer inference traces. Moreover, when such behaviors are learned mainly from outcome reward, the resulting intermediate crops or visual cues can be noisy or fail to faithfully capture task-relevant visual evidence. In this work, we ask whether the reasoning benefits of ``Thinking with Images'' can be internalized through \emph{Thinking with Imagination}: an internal process that decides where to look and imagines what visual cues closer inspection would reveal without actually invoking tools. We propose \textbf{\ours{}}, an on-policy self-distillation framework in which a teacher plays the role of a ``Thinking with Images'' reasoner during training: it receives privileged zoomed evidence views derived from annotated regions, and supervises the model's own imagination reasoning trajectories. \ours{} does not require an external teacher or high-quality imagination demonstrations. Experiments on vision-centric benchmarks show that \ours{} achieves the best average performance among compared models while significantly reducing inference overhead compared with ``Thinking with Images'' methods. Our code is publicly available at~\url{https://github.com/walkeralan123/Imagine-OPD}.
\end{abstract}

\begin{figure*}[t]
\centering
\includegraphics[width=0.85\textwidth]{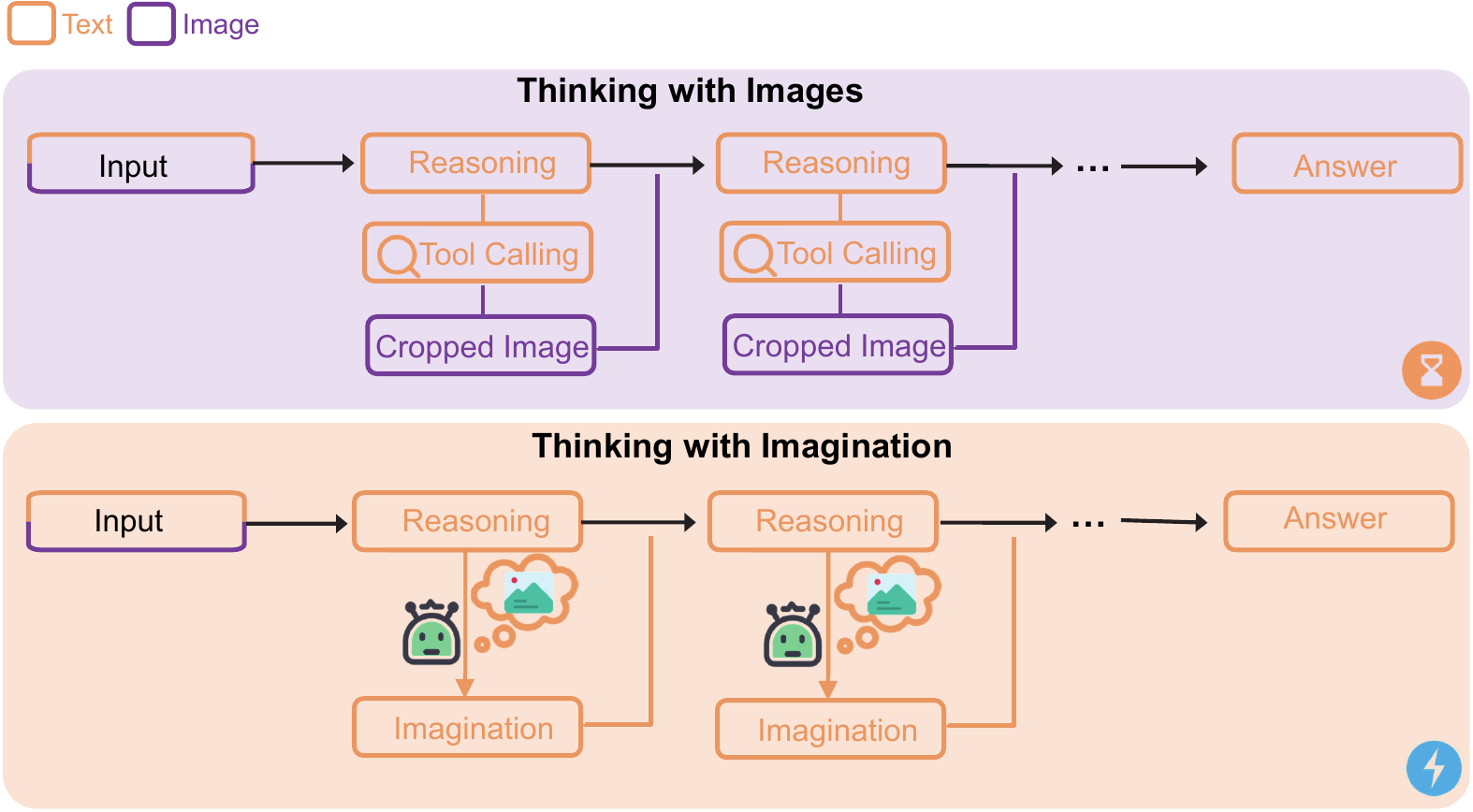}
\caption{\textbf{From ``Thinking with Images'' to ``\textit{Thinking with Imagination}''.} ``Thinking with Images'' interleaves text reasoning with explicit tool calls and cropped images, improving access to local evidence but incurring repeated tool interaction and visual encoding. \textit{Thinking with Imagination} replaces these intermediate images with an internal imagination process that can carry forward the same fine-grained visual evidence without actually invoking tools.}
\label{fig:overview}
\end{figure*}

\section{Introduction}

While Multimodal Large Language Models (MLLMs) perform well on general vision-language tasks~\citep{bai2025qwen25vltechnicalreport, bai2025qwen3vltechnicalreport, openai2024gpt4ocard}, they often fail to deliver robust performance on fine-grained visual reasoning. In these scenarios, the key visual cues are typically small, localized, and highly susceptible to being diluted by the surrounding background~\citep{wu2023vguidedvisualsearch, fu2024blinkmultimodallargelanguage, zhang2025mllmsknowlooktrainingfree, wang2024divideconquercombinetrainingfree, zhang2025mmerealworldmultimodalllmchallenge, wei2026zoomingzoomingregiontoimagedistillation}. During inference, the model must identify these localized clues directly from the entire image. This forces critical target details such as low-resolution characters, fine-grained properties, or complex object relations to compete with massive background tokens, thereby hindering accurate distinction in cluttered scenes~\citep{wu2023vguidedvisualsearch, fu2024blinkmultimodallargelanguage, zhang2025mllmsknowlooktrainingfree, wang2026traceableevidenceenhancedvisual, wei2026zoomingzoomingregiontoimagedistillation, yuan2026visionopdlearningfinedetails}.

To address this limitation, recent ``Thinking with Images'' methods empower MLLMs with explicit image manipulation tools during reasoning, such as zooming into local regions and reasoning over the resulting intermediate images~\citep{su2025thinkingimagesmultimodalreasoning, wu2023vguidedvisualsearch, hu2024visualsketchpadsketchingvisual, fu2025refocusvisualeditingchain, zheng2026deepeyesincentivizingthinkingimages, wang2025pixelreasonerincentivizingpixelspace, zhang2025thymethinkimages, wu2026vtoolr1vlmslearnthink}. This paradigm is effective because it gives the model direct access to isolated local evidence that matters for the question. However, it also introduces two notable limitations: (1) it requires redundant tool invocations, leading to lengthy inference traces and substantially reduced efficiency~\citep{hu2024visualsketchpadsketchingvisual, zheng2026deepeyesincentivizingthinkingimages, zhang2025thymethinkimages, wei2026zoomingzoomingregiontoimagedistillation}; and (2) when such behaviors are learned mainly from outcome-level reward~\citep{zhang2025thymethinkimages, zheng2026deepeyesincentivizingthinkingimages, wang2025pixelreasonerincentivizingpixelspace, wu2026vtoolr1vlmslearnthink, wang2026traceableevidenceenhancedvisual}, the generated intermediate crops or visual cues may be noisy or fail to faithfully capture task-relevant visual evidence~\citep{liu2025faithfulnessvisualthinkingmeasurement, li2026reliablethinkingimages}.

When humans comprehend complex visual scenes, they often do not need to physically zoom in or crop the image. Instead, they mentally direct their attention to different visual regions, imagining what closer inspection would reveal, and use this imagined evidence to support reasoning. Inspired by this insight, we ask: \textit{Can we distill the advantages of ``Thinking with Images'' via training, thereby enabling the model to perform reasoning without invoking tools during inference?}

To answer this question, we propose \textit{Thinking with Imagination}, an internalized paradigm that replaces explicit visual manipulation with purely textual ``implicit imagination.'' In this formulation, the model performs an internal imagination process at each reasoning step: it first decides where to look and then imagines what visual evidence closer inspection of that region would reveal. The entire reasoning trajectory remains textual: each imagination step consists of a region selection and a local observation description, enabling single-pass inference from the original image without any tool calls.

A natural follow-up question is how to train such imagination capability. Two intuitive approaches each have notable shortcomings. Supervised fine-tuning (SFT) requires high-quality imagination demonstrations and trains off-policy, creating a mismatch between the training data distribution and the model's own generation behavior~\citep{gu2026minillmonpolicydistillationlarge, agarwal2024onpolicydistillationlanguagemodels}. Reinforcement learning with verifiable reward (RLVR) supplies outcome-level reward signals without directly supervising the quality of the imagined content itself~\citep{zheng2026deepeyesincentivizingthinkingimages, zhang2025thymethinkimages, wu2026vtoolr1vlmslearnthink, yuan2026visionopdlearningfinedetails}. We therefore propose \textbf{Imagine-OPD}, an on-policy self-distillation framework that combines the benefits of on-policy training with dense process-level supervision.

Concretely, Imagine-OPD instantiates a ``Thinking with Images'' teacher and a student that reasons from the original image. Initially, the student samples on-policy imagination trajectories conditioned on the input image and question. For each resulting prefix, a teacher initialized from the same base model conditions on privileged zoomed evidence and provides token-level guidance through a KL distillation objective~\citep{zhao2026selfdistilledreasoneronpolicyselfdistillation, zhang2026fasteffectiveonpolicydistillation, fu2026revisitingonpolicydistillationempirical}. In this way, the student learns to align its imagination trajectories with the local evidence available to the teacher, while requiring only the global image at inference time. Notably, Imagine-OPD does not require an external teacher model or pre-constructed high-quality imagination demonstrations; the supervision signal comes entirely from privileged visual crops conditioning on the same base model.

We evaluate Imagine-OPD on four vision-centric benchmarks. Imagine-OPD consistently
  improves both 4B and 8B backbones across all benchmarks and achieves the best average score among compared
  proprietary models, open-source MLLMs, and explicit ``Thinking with Images'' methods. It also runs 1.5--2.7$
  \times$ faster than tool-augmented baselines by using only a single original-image inference pass. Attention
  analysis further shows increased coverage on annotated evidence regions, indicating that the model better
  internalizes where to inspect during visual reasoning.

Our contributions are as follows:
\begin{itemize}
    \item We formulate \emph{Thinking with Imagination} as an internalized paradigm of ``Thinking with Images'', where the model reasons from the original image by deciding where to look, imagining what closer inspection would reveal, and using the imagined local evidence for subsequent reasoning.
    \item We propose \ours{}, an on-policy self-distillation framework in which a ``Thinking with Images'' teacher reasons with zoomed evidence views and supervises the model's own imagination trajectories, enabling dense process supervision without requiring high-quality imagination demonstrations.
    \item Comprehensive experiments show that \ours{} yields strong fine-grained visual reasoning performance across vision-centric benchmarks while substantially reducing the inference overhead compared with ``Thinking with Images'' methods.
\end{itemize}

\begin{figure*}[t]
\centering
\includegraphics[width=0.85\textwidth]{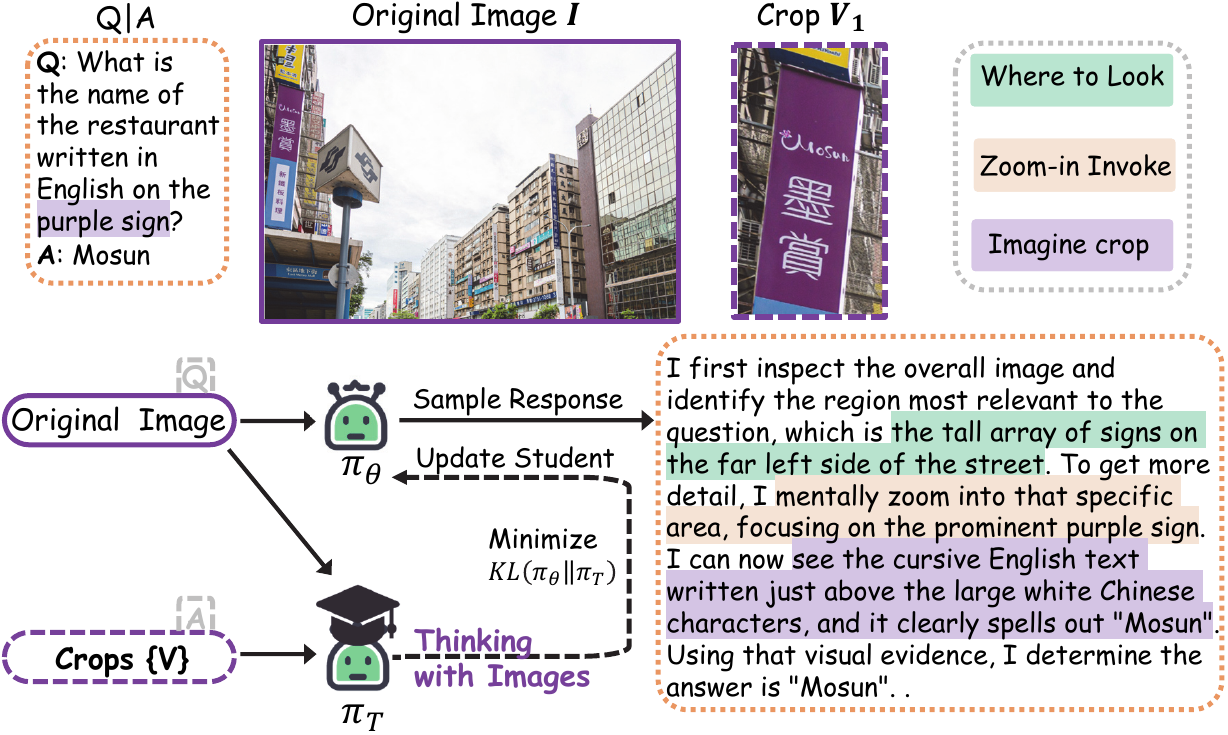}
\caption{\textbf{A concrete training instance of \ours{}.} Given an original image and a question, the student model generates its own imagination trace. A teacher then evaluates the same trajectory with privileged cropped evidence views and answer, providing on-policy supervision through a token-level KL objective.}
\label{fig:method-detail}
\end{figure*}

\section{Related Work}

\paragraph{Thinking with Images.}
Recent multimodal reasoning methods increasingly treat images not only as static inputs, but also as external workspaces that can be actively manipulated during inference~\citep{su2025thinkingimagesmultimodalreasoning}. In this ``Thinking with Images'' paradigm, a model interleaves textual reasoning with visual actions: it decides where to inspect, invokes an operation such as zooming, cropping, drawing, or code-based image editing, and then reasons over the resulting intermediate visual observation~\citep{wu2023vguidedvisualsearch, hu2024visualsketchpadsketchingvisual, fu2025refocusvisualeditingchain, zheng2026deepeyesincentivizingthinkingimages, zhang2025thymethinkimages, wang2025pixelreasonerincentivizingpixelspace, wu2026vtoolr1vlmslearnthink}. This paradigm is effective for fine-grained visual reasoning because it makes small, localized evidence directly visible and reduces interference from irrelevant global context. However, when inference-time visual actions are learned mainly from outcome-level reward, the generated crops or visual cues may be noisy, insufficient, or weakly connected to the evidence needed for the answer~\citep{liu2025faithfulnessvisualthinkingmeasurement, li2026reliablethinkingimages, du2025revisitingnecessitylengthychainofthought}. Our work keeps the central insight of Thinking with Images, namely that reasoning should be guided by local visual evidence, but removes explicit image manipulation at inference time by training the model to internalize the inspection process as textual imagination.

\paragraph{Visual Imagination and Latent Visual Reasoning.}
Visual imagination studies whether multimodal models can construct intermediate visual content during reasoning, rather than only reasoning over the image representation given at the input. Some methods make this imagination explicit by generating or editing visual states, such as sketches, images, or visual workspaces, which can then be inspected in later reasoning steps~\citep{hu2024visualsketchpadsketchingvisual, li2025imaginereasoningspacemultimodal}. Other methods move imagination into latent space, using hidden states or latent visual tokens to represent imagined visual content without rendering it as pixels or text~\citep{li2025latentvisualreasoning, yang2025machinementalimageryempower, wang2025monetreasoninglatentvisual}. While latent-space imagination avoids explicit visual generation, recent analysis suggests that such latent states may be difficult to interpret and not always faithfully connected to the input image or the final answer~\citep{li2026imaginationhelpsvisualreasoning}. Our work takes a text-space route to visual imagination, making the intermediate process directly readable. More importantly, we do not require high-quality imagination trajectories as supervision.

\paragraph{On-Policy Distillation and Self-Distillation.}
Prior work has shown that on-policy distillation can reduce the mismatch between teacher-guided training and autoregressive inference by supervising models on their own generated trajectories~\citep{gu2026minillmonpolicydistillationlarge, lu2025onpolicydistillation, agarwal2024onpolicydistillationlanguagemodels, zhang2026fasteffectiveonpolicydistillation, fu2026revisitingonpolicydistillationempirical}. More recent on-policy self-distillation methods further show that token-level supervision can be transferred from the model itself when the teacher is given privileged context, such as stronger demonstrations, ground-truth solutions, or richer feedback~\citep{zhao2026selfdistilledreasoneronpolicyselfdistillation, shenfeld2026selfdistillationenablescontinuallearning, hübotter2026reinforcementlearningselfdistillation}. Concurrently, Vision-OPD applies on-policy self-distillation to fine-grained multimodal perception by using a crop-conditioned teacher to supervise a full-image student~\citep{yuan2026visionopdlearningfinedetails}. \ours{} shares the idea of privileged regional supervision, but uses it to learn \textit{Thinking with Imagination}, enabling it to decide where to inspect and imagine the relevant local evidence without test-time crops or tools.

\section{Thinking with Imagination}

\subsection{From Thinking with Images to Thinking with Imagination}

Thinking with Images improves fine-grained visual reasoning by making visual inspection an explicit part of the reasoning process. Given an image-question pair $(I,Q)$, a model interleaves textual reasoning with visual manipulations: at step $k$, it selects an operation $a_k$, such as cropping or zooming into a region, and obtains an intermediate image $I'_k=\mathcal{T}(I,a_k)$~\citep{wu2023vguidedvisualsearch, zheng2026deepeyesincentivizingthinkingimages, zhang2025thymethinkimages}. This process can be distilled into two critical steps: deciding where to inspect and obtaining the local evidence revealed by that inspection.

Humans can also reason about complex scenes by mentally focusing on relevant regions and imagining what closer inspection would reveal, but without physically cropping or manipulating the image. Motivated by this ability, we seek to elicit a similar internal inspection process in MLLMs. \textit{Thinking with Imagination} implements this idea by replacing each explicit inspection step $(a_k,I'_k)$ with a textual imagination step $z_k=(\ell_k,o_k)$, where $\ell_k$ specifies the region to mentally inspect and $o_k$ imagines what closer inspection of that region would reveal. In this formulation, $\ell_k$ plays the role of the inspection decision $a_k$, while $o_k$ plays the role of the local observation obtained from $I'_k$. Figure~\ref{fig:overview} summarizes this shift from explicit image manipulation to internal textual imagination.

The resulting reasoning trace interleaves textual reasoning with imagined visual evidence. Let $z_{1:K}=(z_1,\dots,z_K)$ denote the imagination trajectory, where each step can introduce a local text fragment, subtle attribute, object relation, or spatial cue revealed by imagined closer inspection.

\subsection{\ours{}: Thinking with Imagination via On-Policy Self-Distillation}

Learning Thinking with Imagination requires supervision for both parts of each imagination step: the selected inspection target and the imagined local observation. We use zoomed evidence views as the privileged visual context for this supervision. Each training example contains an image-question pair $(I,Q)$, annotated evidence boxes $B_{1:M}=(B_1,\dots,B_M)$, and an answer $A$. From these boxes, we derive zoomed evidence views $I'_{1:M}=\{\mathrm{crop}(I,B_m)\}_{m=1}^{M}$. The student generates an imagination trajectory from $(I,Q)$, while the teacher plays the role of a Thinking-with-Images reasoner by evaluating the same trajectory with access to $(I,Q,I'_{1:M},A)$. The teacher is initialized from the same base model as the student and kept fixed during training.

We train the student on its own generated trajectories, following the on-policy distillation principle~\citep{gu2026minillmonpolicydistillationlarge, agarwal2024onpolicydistillationlanguagemodels, zhao2026selfdistilledreasoneronpolicyselfdistillation}. Let $y=(z_{1:K},\hat{A})\sim\pi_\theta(\cdot\mid I,Q)$ denote a student-generated output. At each token position $t$, the student distribution is $q_t(\cdot)=\pi_\theta(\cdot\mid I,Q,y_{<t})$, while the teacher distribution is $p_t(\cdot)=\pi_{\bar{\theta}}(\cdot\mid I,Q,I'_{1:M},A,y_{<t})$. We minimize the averaged token-level reverse KL along the student rollout:
\begin{equation}
\mathcal{L}_{\text{distill}}(\theta)
=
\mathbb{E}
\left[
\frac{1}{|y|}\sum_{t=1}^{|y|}
D_{\mathrm{KL}}(q_t\|p_t)
\right],
\end{equation}
where the expectation is over training examples and student-generated trajectories. Because both distributions are evaluated on the student's own prefixes, the teacher supervises the decisions the model actually makes, including partially correct or drifting imagination steps. This objective encourages the student to assign higher probability to reasoning continuations that are compatible with the teacher's zoomed evidence views, while reducing continuations that are weakly supported by the relevant local observations.

This design uses intermediate images only during training. The teacher can judge the student's imagination with privileged local evidence, but gradients are applied to the student that condition on the original image and question. Compared with supervised fine-tuning on fixed imagination traces, the supervision is applied on the student's own prefixes; compared with outcome-level reinforcement learning, every token receives dense evidence-conditioned guidance. Figure~\ref{fig:method-detail} illustrates this teacher-student alignment on a concrete training instance. At inference time, the model receives only the original image and question and performs visual reasoning through \textit{Thinking with Imagination}, without explicit image manipulation.

\begin{table*}[t]
\centering
\caption{\textbf{Performance comparison across perception-centric and visual reasoning benchmarks.} We report accuracy on V*, HR-Bench-4K, HR-Bench-8K, and MME-RealWorld-Lite. Missing results are marked with ``-''. The best result in each column is boldfaced, and the second-best result is underlined.}
\label{tab:main-results}
\resizebox{\textwidth}{!}{%
\begin{tabular}{lccccccccccccc}
\toprule
\multirow{2}{*}{\textbf{Model}} & \multicolumn{3}{c}{\textbf{V*}} & \multicolumn{3}{c}{\textbf{HR-Bench-4K}} & \multicolumn{3}{c}{\textbf{HR-Bench-8K}} & \multicolumn{3}{c}{\textbf{MME-RealWorld-Lite}} & \multirow{2}{*}{\textbf{Avg.}} \\
\cmidrule(lr){2-4}\cmidrule(lr){5-7}\cmidrule(lr){8-10}\cmidrule(lr){11-13}
& Overall & Attr. & Spa. & Overall & FSP & FCP & Overall & FSP & FCP & Overall & Rea. & Perc. & \\
\midrule\rowcolor[HTML]{b5ded5}
\multicolumn{14}{c}{\textbf{Proprietary Model}} \\
GPT-4o & 67.5 & 72.2 & 60.5 & 65.0 & 66.8 & 63.3 & 59.6 & 60.8 & 58.5 & 52.0 & 48.3 & 54.4 & 61.0 \\
Gemini-2.5-Flash & 72.3 & 77.3 & 64.4 & 77.5 & 81.5 & 74.0 & 73.7 & 75.8 & 71.8 & 50.2 & 49.9 & 50.4 & 68.4 \\
Gemini-2.5-Pro & 79.1 & 86.8 & 68.4 & \textbf{83.9} & 85.5 & \textbf{82.3} & \textbf{81.5} & 83.0 & \textbf{80.0} & \underline{58.3} & \textbf{55.1} & 59.9 & 75.7 \\
Gemini-3-Flash & 72.3 & 64.4 & 77.3 & 77.5 & 81.5 & 74.0 & 73.7 & 75.8 & 71.8 & \textbf{61.2} & 51.3 & \underline{65.8} & 71.2 \\
\midrule\rowcolor[HTML]{f0e2d6}
\multicolumn{14}{c}{\textbf{Open-Source Models}} \\
InternVL3-8B & 70.2 & 67.8 & 73.7 & 70.0 & 78.8 & 61.3 & 69.3 & 78.8 & 59.8 & 48.6 & 44.8 & 51.0 & 64.5 \\
Qwen2.5-VL-7B & 74.3 & 77.4 & 69.7 & 68.0 & 80.3 & 55.8 & 63.8 & 73.8 & 53.8 & 42.3 & 35.9 & 46.5 & 62.1 \\
Qwen3-VL-4B & 80.1 & 80.9 & 78.9 & 78.3 & 87.5 & 69.3 & 72.9 & 79.3 & 66.3 & 50.1 & 44.8 & 53.4 & 70.4 \\
Qwen3-VL-8B & 86.4 & 87.0 & 85.5 & 78.9 & 87.8 & 70.0 & 74.6 & 82.3 & 66.8 & 50.3 & 46.2 & 52.9 & 72.6 \\
Qwen2.5-VL-32B & 81.2 & 77.4 & 86.8 & 73.4 & 87.5 & 59.3 & 70.4 & 82.3 & 58.5 & 46.2 & 39.3 & 50.6 & 67.8 \\
Qwen3-VL-30B-Thinking & 82.2 & 81.7 & 82.9 & 78.5 & 88.5 & 68.5 & 74.2 & 80.3 & 68.3 & 53.2 & 46.3 & 58.0 & 72.0 \\
Qwen3-VL-235B-A22B & 80.6 & 82.9 & 79.1 & 83.0 & 89.0 & \underline{77.0} & \underline{80.4} & 83.0 & \underline{77.3} & 56.5 & 50.7 & 60.2 & 75.1 \\
Kimi-K2.5 (1T) & 85.9 & 87.0 & 84.2 & 81.7 & 87.3 & 76.0 & 75.1 & 82.8 & 67.5 & 53.8 & 49.3 & 56.6 & 74.1 \\
\midrule\rowcolor[HTML]{d2cae3}
\multicolumn{14}{c}{\textbf{Reasoning with Tool Methods}} \\
Pixel-Reasoner & 80.6 & 83.5 & 76.3 & 72.9 & 86.0 & 60.3 & 66.9 & 80.0 & 54.3 & 50.1 & 45.6 & 53.1 & 67.6 \\
Thyme & 82.2 & 83.5 & 80.3 & 77.0 & \underline{91.0} & 63.0 & 72.0 & 72.0 & 57.5 & 55.2 & 49.1 & 59.1 & 71.6 \\
DeepEyes & \underline{90.1} & \underline{91.3} & \textbf{88.2} & 75.1 & \textbf{91.3} & 59.0 & 72.6 & 86.8 & 58.5 & 53.2 & 46.0 & 54.0 & 72.8 \\
TreeVGR-7B & \textbf{91.1} & \textbf{94.0} & \underline{87.0} & 77.1 & 90.3 & 64.0 & 73.1 & 86.5 & 59.6 & 54.9 & 50.7 & \textbf{66.0} & 74.1 \\
\midrule\rowcolor[HTML]{e7a375}
\multicolumn{14}{c}{\textbf{Our Models}} \\
\rowcolor[HTML]{DCEBFF}
\textbf{\ours{}-4B} & 88.0 & 89.6 & 85.5 & 83.3 & 90.3 & 76.3 & 78.6 & \underline{88.0} & 69.3 & 56.9 & 53.2 & 59.3 & \underline{76.7} \\
\hspace{1em}$\Delta$ (vs Qwen3-VL-4B) & +7.9 & +8.7 & +6.6 & +5.0 & +2.8 & +7.0 & +5.7 & +8.7 & +3.0 & +6.8 & +8.4 & +5.9 & +6.3 \\
\rowcolor[HTML]{DCEBFF}
\textbf{\ours{}-8B} & 87.4 & 88.7 & 86.8 & \underline{83.5} & 90.8 & 76.3 & 79.5 & \textbf{88.3} & 70.8 & 58.1 & \underline{54.3} & 60.6 & \textbf{77.1} \\
\hspace{1em}$\Delta$ (vs Qwen3-VL-8B) & +1.0 & +1.7 & +1.3 & +4.6 & +3.0 & +6.3 & +4.9 & +6.0 & +4.0 & +7.8 & +8.1 & +7.7 & +4.5 \\
\bottomrule
\end{tabular}%
}
\end{table*}

\begin{figure*}[t]
\centering
\begin{tcolorbox}[enhanced, colback=white, colframe=black!55, boxrule=0.5pt, arc=1.5mm, left=1mm, right=1mm, top=1mm, bottom=1mm]
\vspace{0.6em}
\begin{minipage}{\textwidth}
    \centering
    \begin{minipage}{0.31\textwidth}
        \centering
        \includegraphics[width=\textwidth]{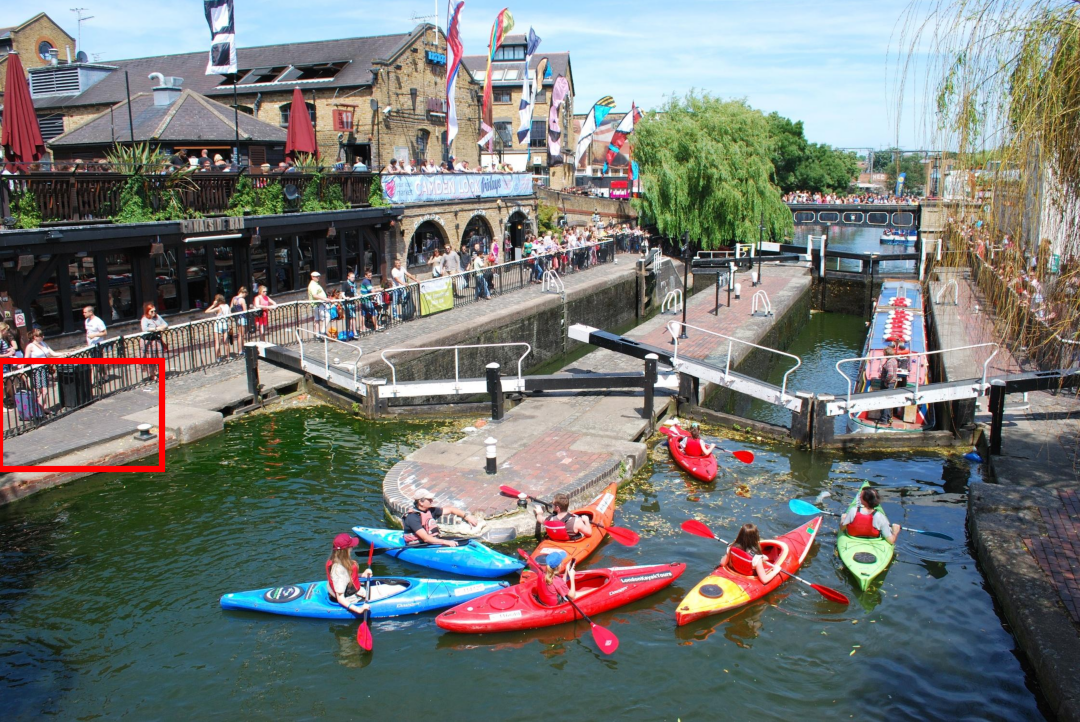}

        {\small \textbf{Original Image}}
    \end{minipage}
    \hfill
    \begin{minipage}{0.31\textwidth}
        \centering
        \includegraphics[width=\textwidth]{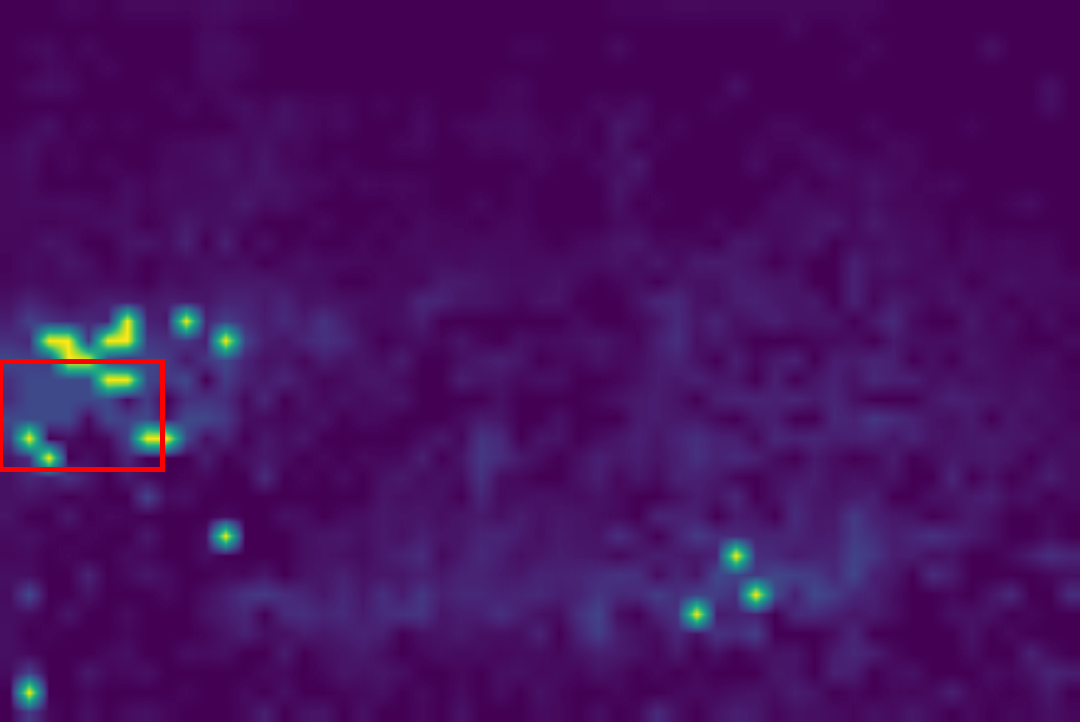}

        {\small \textbf{Qwen3-VL-4B}}
    \end{minipage}
    \hfill
    \begin{minipage}{0.31\textwidth}
        \centering
        \includegraphics[width=\textwidth]{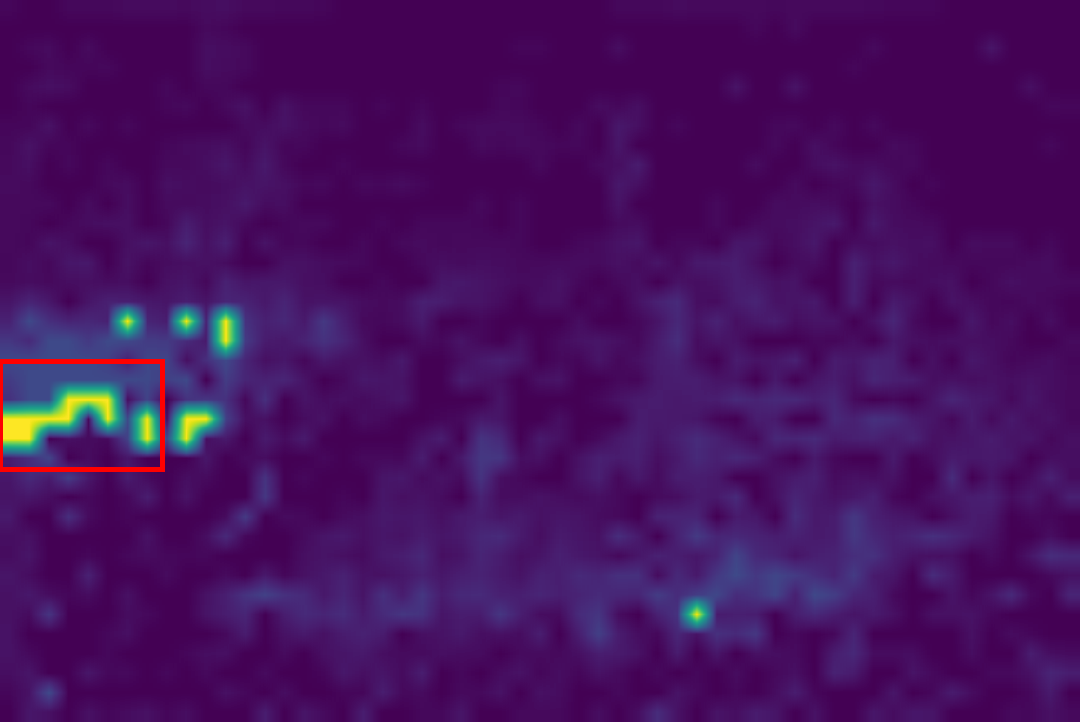}

        {\small \textbf{\ours{}-4B}}
    \end{minipage}
\end{minipage}

\vspace{0.6em}
\hrule
\vspace{0.6em}

\textbf{Question:} Is the suitcase on the left or right side of the river? \textbf{A.} right \textbf{B.} left \textbf{Answer:} B

\textbf{\ours{}-4B Response:} I first inspect the overall image and identify the region most relevant to the question. To get more detail, I mentally zoom into the left side of the riverbank where people are gathered. I can now see a person standing near a black metal railing, holding a suitcase. The suitcase is positioned on the left side of the river, adjacent to the walkway and the buildings. Using that visual evidence, I determine the answer.<answer>B. left</answer>

\end{tcolorbox}
\caption{\textbf{Qualitative attention analysis.} \ours{} places more attention on the decisive visual evidence compared with the base model.}
\label{fig:attention-case}
\end{figure*}

\section{Experiments}

\subsection{Experimental Setup}

\paragraph{Benchmarks.}
We conduct evaluations on four vision centric benchmarks: V*~\citep{wu2023vguidedvisualsearch}, HR-Bench-4K, HR-Bench-8K~\citep{wang2024divideconquercombinetrainingfree}, and MME-RealWorld-Lite~\citep{zhang2025mmerealworldmultimodalllmchallenge}.

\paragraph{Baselines.}
We consider three groups of baselines. First, we report proprietary models, including GPT-4o~\citep{openai2024gpt4ocard}, Gemini-2.5-Flash, Gemini-2.5-Pro~\citep{comanici2025gemini25pushingfrontier}, and Gemini-3-Flash. Second, we compare against open-source MLLMs across a wide parameter range, including InternVL3-8B~\citep{zhu2025internvl3exploringadvancedtraining}, Qwen2.5-VL-7B~\citep{bai2025qwen25vltechnicalreport}, Qwen3-VL-4B, Qwen3-VL-8B, Qwen2.5-VL-32B, Qwen3-VL-30B-Thinking, Qwen3-VL-235B-A22B~\citep{bai2025qwen3vltechnicalreport}, and Kimi-K2.5 (1T)~\citep{kimiteam2026kimik25visualagentic}. Third, we compare against explicit ``Thinking with Images'' methods, including Pixel-Reasoner~\citep{wang2025pixelreasonerincentivizingpixelspace}, Thyme~\citep{zhang2025thymethinkimages}, DeepEyes~\citep{zheng2026deepeyesincentivizingthinkingimages}, and TreeVGR-7B~\citep{wang2026traceableevidenceenhancedvisual}. In addition, we include a GRPO training baseline~\citep{shao2024deepseekmathpushinglimitsmathematical} in the ablation analysis. For GRPO, we use learning rate $1\times10^{-6}$, 8 rollouts per prompt, temperature 1.0, and a rule-based answer-matching reward.

\paragraph{Training Data.}
Our training data are built from the original training set of V*~\citep{wu2023vguidedvisualsearch}. We filter the training set using Qwen3-VL-4B and retain only samples for which the model answers correctly at least 2 times in 4 trials. This procedure yields 19K training samples. Each retained sample is paired with one or more ground-truth bounding boxes, from which we derive the zoomed evidence views used by the teacher during training.

\paragraph{Training Configuration.}
Our model is built on Qwen3-VL-4B and Qwen3-VL-8B~\citep{bai2025qwen3vltechnicalreport}. Training is conducted over 1 epoch with a global batch size of 32 and a learning rate of $1 \times 10^{-6}$ on 4 × A800 GPUs. During generation, we sample one rollout per prompt, setting the temperature parameter to 1.0 and capping the maximum response length at 256 tokens. The detailed prompts used in training and inference are provided in Appendix~\ref{app:prompts}.

\subsection{Main Results}

\textbf{Consistent improvements across benchmarks.}
Table~\ref{tab:main-results} demonstrates that \ours{} consistently improves Qwen3-VL backbones on all four benchmarks. \ours{}-4B improves the average score from 70.4 to 76.7, with gains of +7.9 on V*, +5.0 on HR-Bench-4K, +5.7 on HR-Bench-8K, and +6.8 on MME-RealWorld-Lite. \ours{}-8B also improves Qwen3-VL-8B from 72.6 to 77.1 average score. These results show that on-policy supervision from zoomed evidence views improves both fine-grained perception and real-world visual reasoning.

\noindent\textbf{Strong average performance with small backbones.}
\ours{} achieves the best average score among all compared models while using only 4B/8B backbones. In particular, \ours{}-8B reaches 77.1 average score, higher than Qwen3-VL-30B-Thinking (72.0), Qwen3-VL-235B-A22B (75.1), and Kimi-K2.5 (1T) (74.1). \ours{}-4B also reaches 76.7 average score, outperforming the Qwen3-VL-8B backbone (72.6) and Gemini-2.5-Pro (75.7) on average.

\noindent\textbf{Exceeding explicit ``Thinking with Images'' methods.}
We further demonstrate the clear superiority of \ours{} over ``thinking with images'' methods. \ours{}-8B obtains a 77.1 average score, and \ours{}-4B obtains 76.7, outperforming TreeVGR-7B (74.1), DeepEyes (72.8), Thyme (71.6), and Pixel-Reasoner (67.6). Notably, \ours{} improves over TreeVGR-7B by +6.4 on HR-Bench-4K, +6.4 on HR-Bench-8K, and +3.2 on MME-RealWorld-Lite. This suggests that \ours{} distills the advantage of tool-use into the model, instead of relying on iterative crop operations at test time.

\begin{figure}[t]
\centering
\includegraphics[width=0.95\columnwidth]
{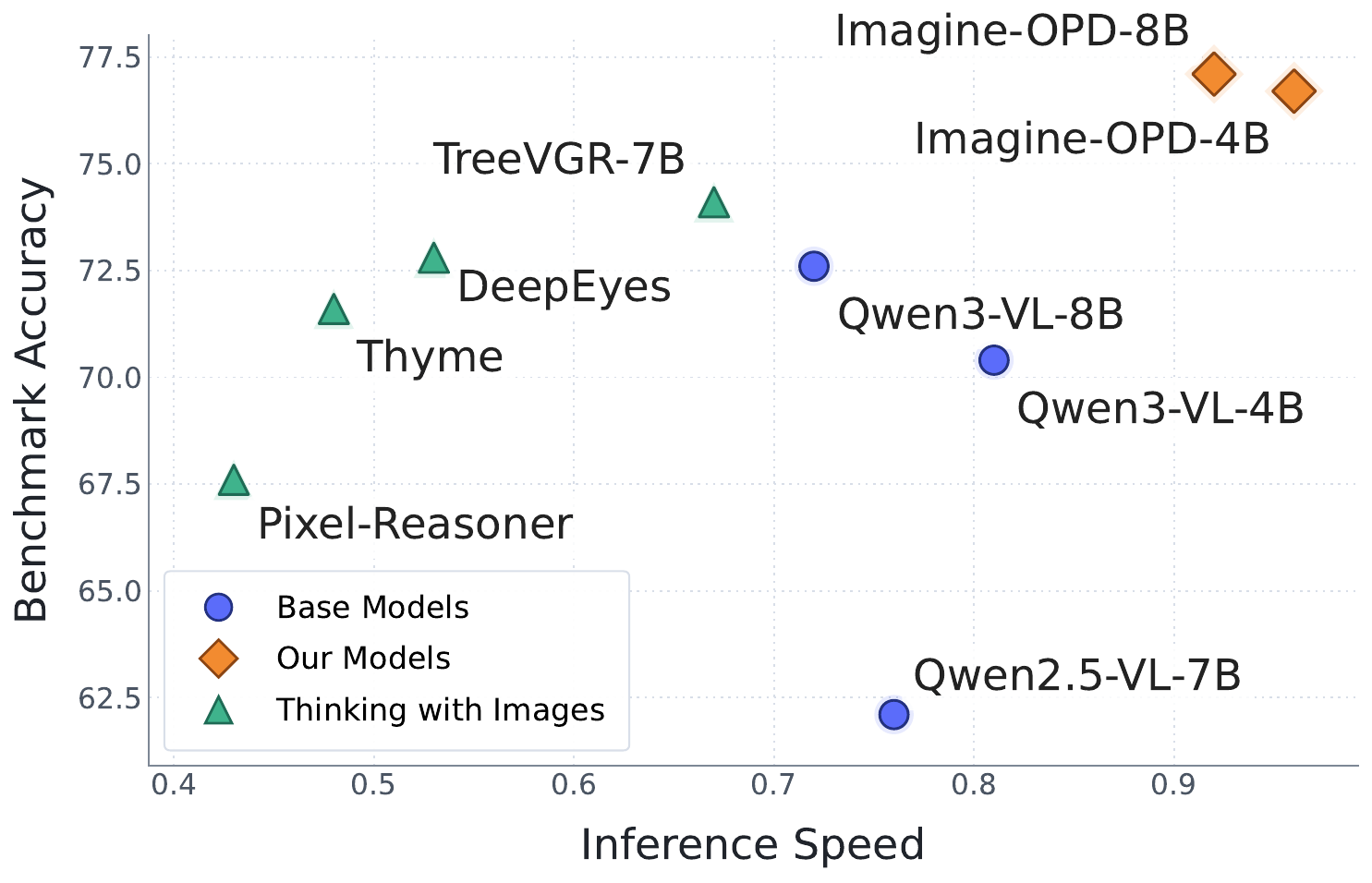}
\caption{Trade-off between benchmark accuracy and inference speed.}
\label{fig:efficiency}
\end{figure}

\begin{table}[t]
\centering
\caption{\textbf{Attention coverage analysis on TreeBench.} We report accuracy and the percentage of relative attention covered by annotated evidence regions on TreeBench~\citep{wang2026traceableevidenceenhancedvisual}.}
\label{tab:attention-coverage}
\begin{tabular}{lcc}
\toprule
\textbf{Model} & \textbf{Acc.} & \textbf{Coverage (\%)} \\
\midrule
Qwen3-VL-4B & 42.7 & 23.6 \\
Qwen3-VL-8B & 45.7 & 26.0 \\
\ours{}-4B & 46.4 & 27.4  \\
\ours{}-8B & \textbf{49.6}  & \textbf{29.6}  \\
\bottomrule
\end{tabular}
\end{table}

\section{Analysis}
\subsection{Efficiency Analysis}

Figure~\ref{fig:efficiency} shows the speed-accuracy trade-off compared with representative Thinking with Image methods, including TreeVGR-7B, DeepEyes, Thyme, and Pixel-Reasoner. Inference speed is measured in samples per second, and benchmark accuracy corresponding to the average score provided in Table~\ref{tab:main-results}. Explicit Thinking-with-Images methods obtain competitive accuracy, but their repeated tool calls and visual encoding make inference speed substantially slower. In contrast, \ours{} keeps inference to a single image pass and lies on the speed-accuracy frontier. \textbf{Concretely, \ours{} is 1.5--2.7$\times$ faster than representative explicit Thinking-with-Images methods, while also improving average benchmark performance. }

\begin{table*}[t]
\centering
\caption{\textbf{Ablation of privileged teacher information.} We vary the privileged context available to the teacher in OPD. The full setting uses answer-conditioned evidence views cropped from ground-truth (GT) boxes; ``Answer only'' removes evidence views, ``w/o answer'' removes answer, and ``Self boxes'' constructs evidence views from model-proposed boxes.}
\label{tab:teacher-info-ablation}
\begin{tabular}{lccc|ccc}
\toprule
\textbf{Method} & \textbf{Evidence Views} & \textbf{Region Source} & \textbf{Answer} & \textbf{V*} & \textbf{HR-4K} & \textbf{HR-8K} \\
\midrule
Qwen3-VL-4B & -- & -- & -- & 80.1 & 78.3 & 72.9 \\
\hspace{0.7em}w/ answer & $\times$ & -- & \checkmark & 82.2 & 78.5 & 73.3 \\
\hspace{0.7em}w/ GT boxes & \checkmark & GT & $\times$ & 87.4 & 79.8 & 75.4 \\
\hspace{0.7em}w/ Self boxes & \checkmark & Self & \checkmark & 86.4 & 79.8 & 75.1 \\
\textbf{\ours{}-4B} & \checkmark & GT & \checkmark & \textbf{88.0} & \textbf{83.3} & \textbf{78.6} \\
\bottomrule
\end{tabular}
\end{table*}

\begin{table}[t]
\centering
\caption{\textbf{Ablation of training strategy and teacher update.} Starting from Qwen3-VL-4B, we compare GRPO with \ours{}. Under \ours{}, the frozen-teacher variant keeps the privileged teacher fixed at the initial checkpoint, while the online-teacher variant uses the current student checkpoint as the privileged teacher at each update.}
\label{tab:training-strategy-ablation}
{\setlength{\tabcolsep}{4pt}
\begin{tabular}{lccc}
\toprule
\textbf{Method} & \textbf{V*} & \textbf{HR-4K} & \textbf{HR-8K} \\
\midrule
Qwen3-VL-4B & 80.1 & 78.3 & 72.9 \\
\hspace{0.7em}+GRPO & 83.8 & 79.5 & 74.3 \\
\rowcolor[HTML]{EFEFEF}
\multicolumn{4}{l}{\textbf{\hspace{0.7em}+\ours{}}} \\
\hspace{1.4em}w/ frozen teacher & \textbf{88.0} & \textbf{83.3} & \textbf{78.6} \\
\hspace{1.4em}w/ online teacher & 0.0 & 0.0 & 0.0 \\
\bottomrule
\end{tabular}
}
\end{table}

\subsection{Attention Analysis}

Table~\ref{tab:attention-coverage} reports attention coverage analysis on TreeBench~\citep{wang2026traceableevidenceenhancedvisual}, a visual grounded reasoning benchmark that provides question-specific evidence bounding boxes. This allows us to evaluate whether the model attends to the image regions that support the answer. We compute question-conditioned relative attention, map each annotated bounding box to the visual-token grid, and measure the fraction of visual-token attention mass that falls inside the evidence region following prior work~\citep{zhang2025mllmsknowlooktrainingfree, wei2026zoomingzoomingregiontoimagedistillation}. Compared with the corresponding Qwen3-VL backbones, \ours{} improves both accuracy and evidence region coverage: \ours{}-4B improves coverage from 23.6 to 27.4, and \ours{}-8B improves it from 26.0 to 29.6. This shows that the gains are accompanied by more accurate visual attention allocation, consistent with the goal of internalizing where to look. Figure~\ref{fig:attention-case} shows a representative case where \ours{} assigns more relative attention to the decisive region and produces an imagination trajectory grounded in the corresponding region.

\subsection{Ablation Study}

Tables~\ref{tab:teacher-info-ablation} and~\ref{tab:training-strategy-ablation} isolate the supervision source and training strategy of \ours{}. \textbf{(1) Zoomed evidence views provide the main teacher signal.} In Table~\ref{tab:teacher-info-ablation}, removing evidence views and keeping only answer conditioning drops V* from 88.0 to 82.2, HR-4K from 83.3 to 78.5, and HR-8K from 78.6 to 73.3. Removing the answer hint is less harmful, indicating that local visual evidence is more important than answer awareness for supervising imagination. Replacing GT boxes with self-proposed boxes recovers part of the gain but remains weaker than using annotated evidence, showing that region quality matters when constructing the teacher's privileged views. \textbf{(2) Dense OPD supervision is more effective than outcome reward, but requires a stable teacher policy.} Table~\ref{tab:training-strategy-ablation} shows that GRPO improves over the backbone, but remains below OPD with a frozen teacher, suggesting that sparse outcome-level reward is insufficient for learning high-quality imagination trajectories. When the privileged teacher instead uses the current student checkpoint at each update, performance collapses to 0.0 accuracy on all benchmarks. This indicates that the teacher's distribution must remain a stable reference when transferring privileged zoomed evidence into the student's textual imagination process.

\section{Conclusion}

We present \textbf{\ours{}}, an on-policy self-distillation framework that transfers the benefit of ``Thinking with Images'' into \textit{Thinking with Imagination}. By training a teacher with zoomed evidence views to supervise the model's own imagination trajectories, \ours{} learns both where to focus and what visual cues should become available under closer inspection without invoking external tools. Experiments show that \ours{} substantially improves fine-grained visual reasoning and delivers large efficiency gains over tool-augmented visual reasoning, suggesting that the tool-augmented visual reasoning process can be effectively internalized.

\section*{Limitations}

Our current study focuses on zooming-based intermediate views derived from annotated evidence regions. This keeps the supervision signal clear and stable, but also limits the scope of the current conclusions. More diverse and complex image manipulations may require different supervision mechanisms and may not transfer equally well into Thinking with Imagination. In addition, our current experiments require region annotations available during training, extending the framework to weaker or noisier region supervision is an important direction for future work.

\bibliography{main}

\clearpage

\appendix

\section{Relative Attention Map Computation}
\label{app:attention}

Following~\citet{zhang2025mllmsknowlooktrainingfree, wei2026zoomingzoomingregiontoimagedistillation} on relative attention analysis, we visualize which image regions are preferentially used for answering a question, rather than which regions receive high attention in a generic image-understanding setting. Given an image-question pair $(I,Q)$, we first run the model on the target prompt and collect the answer-to-image attention map:
\[
A^{\text{task}}=\mathrm{Attn}(I,Q).
\]
To suppress question-agnostic saliency, we additionally compute a reference attention map under a generic image-description prompt $Q^{\text{ref}}$:
\[
A^{\text{ref}}=\mathrm{Attn}(I,Q^{\text{ref}}).
\]
We then form the relative attention map by subtracting the reference map and renormalizing:
\[
A^{\text{rel}}=\mathrm{Norm}\!\left(A^{\text{task}}-A^{\text{ref}}\right).
\]
In practice, $\mathrm{Attn}(\cdot)$ denotes the aggregated attention from generated answer tokens to visual tokens, averaged across the selected layers and heads, and $\mathrm{Norm}(\cdot)$ rescales the map to $[0,1]$ for visualization. This procedure emphasizes image regions whose contribution increases specifically for the target question.

For the quantitative coverage metric in Table~\ref{tab:attention-coverage}, let $\Omega$ denote the set of visual-token positions covered by the annotated evidence box after mapping the box from image coordinates to the visual-token grid. We keep the positive part of the relative attention map, $\tilde{A}^{\text{rel}}=\max(A^{\text{task}}-A^{\text{ref}},0)$, and compute
\[
\mathrm{Coverage}=100\cdot
\frac{\sum_{j\in\Omega}\tilde{A}^{\text{rel}}_j}
{\sum_j \tilde{A}^{\text{rel}}_j}.
\]
Higher coverage indicates that a larger share of the model's question-specific visual attention falls on the traceable evidence region.

\newpage
\section{Prompts}
\label{app:prompts}

\textbf{Teacher Prompt.}~~
The prompt used for the teacher model is shown below. The teacher receives the original image together with intermediate evidence images and is instructed to reason in text-imagine mode only.

\begin{tcolorbox}[colframe=black!75, arc=2mm,breakable]
\textbf{Instruction.}\\
Solve the visual question step by step in \texttt{text\_imagine} mode only.

\vspace{0.2em}
\textbf{Reasoning Rule.}\\
You will receive the original question and original image(s), along with additional context, including intermediate images produced by applying operations such as cropping and zooming to the original image.

Use this additional context only as hidden support for visual reasoning. You must restate useful operations, such as cropping or zooming as mental image manipulations.

Image sequence:
Image 1 is the original image.
Image 2 and any subsequent images are intermediate images.
Intermediate images produced by applying operations to the original image: zoom.
After knowing which region to focus on, try to solve the problem using only Image 1. Do not mention Image 2, the second image, the intermediate image, or any extra image explicitly.

\vspace{0.2em}
\textbf{Example Format.}\\
I first inspect the overall image and identify the region most relevant to the question. To get more detail, I mentally zoom into the relevant region after the imagined close inspection. I can now see that the sign is triangular and red, which matches only one of the options. Using that visual evidence, I determine the answer.<answer>Your final answer goes here.</answer>

\vspace{0.2em}
\textbf{Output Format (strict adherence required).}
\begin{enumerate}[leftmargin=1.8em, itemsep=0.15em, topsep=0.25em]
    \item When using imagine, describe both the mental operation and what becomes clearer after it.
    \item End with \texttt{<answer>...</answer>}.
    \item If this is a multiple-choice question and your reasoning conflicts with the options, select the single most plausible option once and finish immediately.
\end{enumerate}

\vspace{0.2em}
\textbf{Input Template.}\\
\texttt{<image><image>...}

\vspace{0.2em}
\textbf{Question.}\\
\texttt{\{question\}}

\vspace{0.2em}
\textbf{Ground truth.}\\
\textbf{\{answer\}}
\end{tcolorbox}

\textbf{Student Prompt.}~~
The prompt used for the student model is shown below. In contrast to the teacher, the model receives only the original image and question.

\begin{tcolorbox}[colframe=black!75, arc=2mm]
\textbf{Instruction.}\\
Solve the visual question step by step.

\vspace{0.4em}
\textbf{Reasoning Rule.}\\
When closer inspection is needed, you may use imagination for internal visual reasoning, such as mentally cropping or zooming the image. Describe this as reasoning rather than tool use.

\vspace{0.4em}
\textbf{Example Format.}\\
I first inspect the overall image and identify the region most relevant to the question. To get more detail, I mentally zoom into the relevant region after the imagined close inspection. I can now see that the sign is triangular and red, which matches only one of the options. Using that visual evidence, I determine the answer.<answer>Your final answer goes here.</answer>

\vspace{0.4em}
\textbf{Output Format (strict adherence required).}
\begin{enumerate}[leftmargin=1.8em, itemsep=0.15em, topsep=0.25em]
    \item When using imagine, describe both the mental operation and what becomes clearer after it.
    \item End with \texttt{<answer>...</answer>}.
    \item If this is a multiple-choice question and your reasoning conflicts with the options, select the single most plausible option once and finish immediately.
\end{enumerate}

\vspace{0.4em}
\textbf{Input Template.}\\
\texttt{<image>}

\vspace{0.25em}
\textbf{User's Question.}\\
\texttt{\{question\}}
\end{tcolorbox}

\textbf{Self-Bbox Proposal Prompt.}~~
For the self-proposed bounding-box ablation, we first ask the model to predict a single bounding box from the original image and question, then crop the corresponding intermediate view from the original image, and finally use these model-proposed intermediate views during training in place of the ground-truth evidence crops.

\begin{tcolorbox}[colframe=black!75, arc=2mm]
\textbf{Instruction.}\\
You are selecting a single intermediate crop for visual question answering.

\vspace{0.4em}
\textbf{Goal.}\\
Given the image and the question, identify the smallest single rectangular region in the original image that contains the visual evidence needed to answer the question.

\vspace{0.4em}
\textbf{Rules.}
\begin{enumerate}[leftmargin=1.8em, itemsep=0.15em, topsep=0.25em]
    \item If one object is sufficient, return a tight box around that object.
    \item If the answer depends on multiple objects or their relationship, return one box covering all necessary evidence.
    \item Use integer pixel coordinates on the original image of size \texttt{\{width\}x\{height\}}.
    \item Do not answer the question.
    \item Return exactly one bbox in this format and nothing else: \texttt{<bbox>[x1, y1, x2, y2]</bbox>}.
    \item Always provide your best bbox guess even if the target is small or hard to see. Do not say that the object is absent.
\end{enumerate}

\vspace{0.4em}
\textbf{Question.}\\
\texttt{\{question\}}
\end{tcolorbox}

\end{document}